\documentclass[10pt,twocolumn,letterpaper]{article}

\usepackage[pagenumbers]{cvpr} 

\usepackage{booktabs}   
\usepackage{pifont}     
\newcommand{\cmark}{\ding{51}} 
\newcommand{\xmark}{\ding{55}} 
\usepackage{multirow}
\usepackage{multicol}

\definecolor{cvprblue}{rgb}{0.21,0.49,0.74}
\usepackage[pagebackref,breaklinks,colorlinks,allcolors=cvprblue]{hyperref}

\def\paperID{11138} 
\def\confName{CVPR}
\def\confYear{2026}

\title{CountCluster: Training-Free Object Quantity Guidance with Cross-Attention Map Clustering for Text-to-Image Generation}

\makeatletter
\newcommand{\thanksmark}[1]{\textsuperscript{\@fnsymbol{#1}}}
\makeatother

\author{
    Joohyeon Lee\thanks{Equal contribution} \qquad
    Jin-Seop Lee\thanksmark{1} \qquad
    Jee-Hyong Lee\thanks{Corresponding author} \\
    Sungkyunkwan University, Suwon, South Korea \\
    {\tt\small \{leeju001, wlstjq0602, john\}@skku.edu}
}

\usepackage{comment}
\usepackage{algorithm}
\usepackage{algorithmic}

\usepackage{kotex}

\begin{document}
\maketitle
\begin{abstract}
Text-to-image diffusion models have achieved substantial advances in generating visual content. However, these models often fail to produce the intended number of objects specified in a prompt. Several studies tried to address this limitation by incorporating external counting tools or designing additional object-specific control mechanisms, but these approaches still struggle with consistent quantity alignment and introduce extra computational overhead.
To address this problem without relying on additional training or external modules, we propose CountCluster, a simple and effective method for accurate object-count control. We observe that the object count is strongly influenced by the object cross-attention map (CAM) at the first denoising timestep. Based on this observation, our method aligns early-timestep CAMs with the intended object count. Our method extracts high-activation regions from the object CAM, clusters them into the desired number of groups, and constructs a continuous CAM-aware target map using Gaussian smoothing. By minimizing the KL divergence between the CAM and the target object map at the first denoising timestep, the model is guided to form activation regions corresponding to the specified object count.
Our method enables the model to reflect the intended object count across a wide range of complex and diverse prompt settings while preserving high image quality, all without training or reliance on external modules. Code will be released at https://github.com/JoohyeonL22/CountCluster
\end{abstract}    
\begin{figure}[t]
\centering
\includegraphics[width=1.0\linewidth]{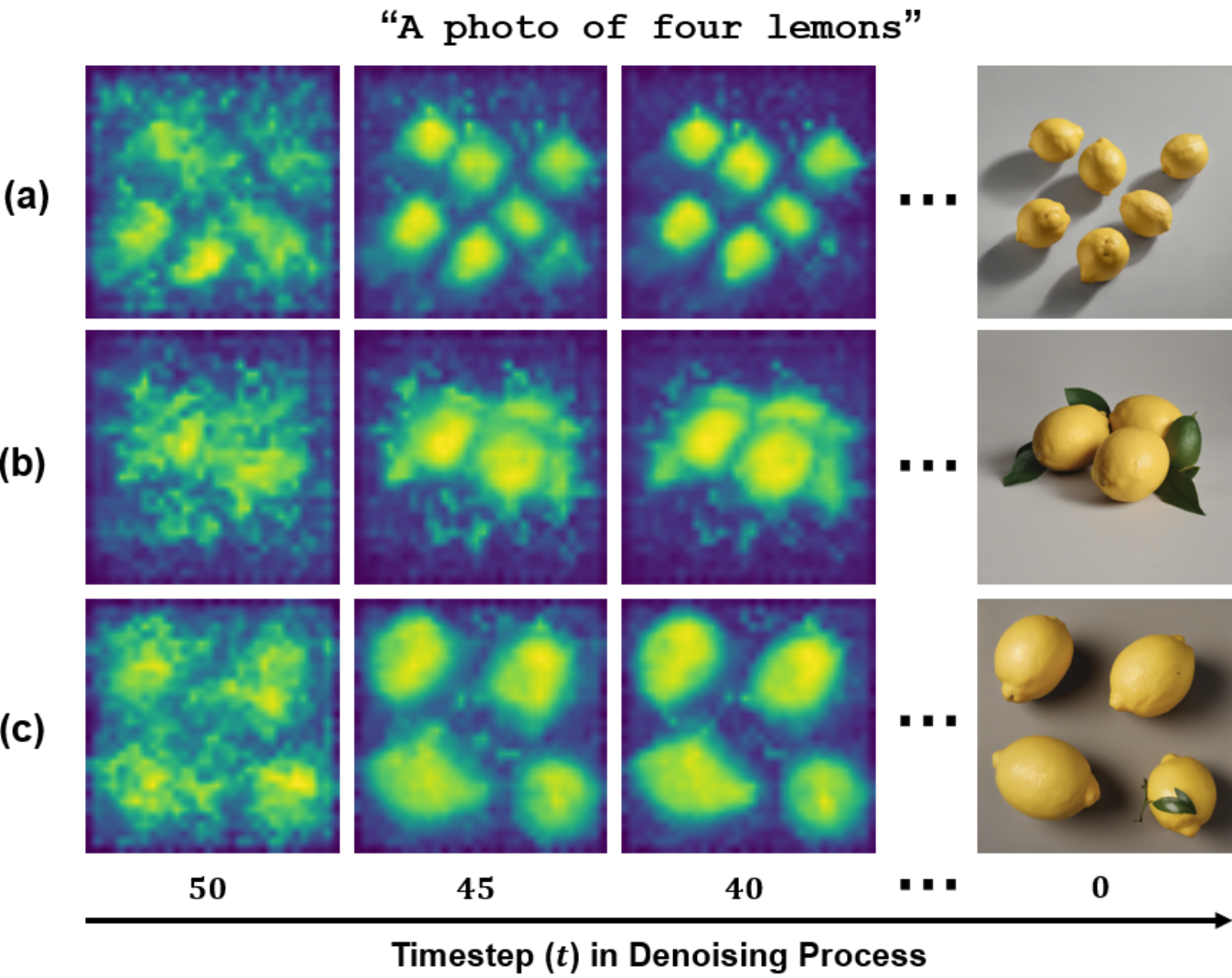}
\caption{
Generated images using SDXL \cite{podell2024sdxl} for the prompt \textit{``A photo of four lemons''}. Each example, generated under the same conditions with different seeds, shows the cross-attention maps of ``lemons'' at timesteps $t=50, 45, 40$ together with the final image at $t=0$. (a) and (b) are failure cases with incorrect instance counts, while (c) is a successful case with the correct number of instances. It can be observed that the positions and number of instances are mostly determined at the early timesteps of the denoising process.
}
\label{figure1}
\vspace{-0.3cm}
\end{figure}

\section{Introduction}
Generative models have led to remarkable progress in the field of text-to-image generation, where visual content is synthesized solely from natural language descriptions. While early approaches were based on GANs \cite{NIPS2014_f033ed80,radford2016unsupervisedrepresentationlearningdeep,Karras_2019_CVPR}, diffusion models \cite{NEURIPS2020_4c5bcfec,10.5555/3540261.3540933,Rombach_2022_CVPR} have recently demonstrated superior performance in terms of image quality and diversity. With the development of diffusion models, various applications including image synthesis \cite{Gu_2022_CVPR,NEURIPS2022_ec795aea,pmlr-v139-ramesh21a}, inpainting \cite{meng2022sdedit,Lugmayr_2022_CVPR,pmlr-v162-nichol22a,couairon2023diffedit}, and visual editing \cite{kawar2023imagic,Tumanyan_2023_CVPR,Cao_2023_ICCV} have been increasingly adopted in practical scenarios. This trend highlights a growing need to generate images that faithfully reflect the user's intent, including the specified number of instances. However, diffusion models still struggle to generate images that correctly reflect the exact number of instances of each object described in the input text (e.g., {\textit ``a photo of four lemons''}) \cite{Paiss_2023_ICCV,Hu_2023_ICCV,wen2023improvingcompositionaltexttoimagegeneration}.

Recently, several studies have been conducted to address the limitations of reflecting object quantity in generated images. Existing approaches can be  categorized into two types: iterative refinement-based methods and quantity-representation guided methods.
Iterative refinement-based approaches \cite{binyamin2024count,kang2025counting} estimate the number of object instances during the image generation process, and then either regenerate the image or iteratively refine it while re-estimating the count.
Quantity-representation guided approaches \cite{zafar2024iterativeobjectcountoptimization,zhang2023zeroshot} encode quantity information either by defining new tokens that learn quantity representations or by extracting latent vectors that control the number of instances. Then, they utilize these quantity representations to guide the generation of other object categories with the same count.
These approaches show improvements over a naive Stable Diffusion \cite{Rombach_2022_CVPR,podell2024sdxl}, however, they still struggle to generate images with the exact number of objects. Importantly, they overlook the most critical factor for quantity control---the number of instances in the generated image is mostly determined during the early timesteps of the diffusion process.

For diffusion models, highly activated regions in the object cross-attention map (CAM), which indicate strong attention to the corresponding object token, are strongly correlated with the object locations in the generated image \cite{hertz2023prompttoprompt}.
As shown in the cross-attention maps for ``lemons'' token in \cref{figure1}, the positions and number of objects are largely determined at the early timesteps of the generation process. 
When the number of clusters of highly activated regions in the object cross-attention map at early timesteps exceeds the object quantity specified in the input prompt, the generated image tends to contain more objects, as shown in \cref{figure1}(a). In contrast, when the boundaries between activated regions are ambiguous and overlapping, some objects may be merged during the generation process, resulting in fewer object instances being generated, as shown in \cref{figure1}(b). Therefore, to ensure that the generated image reflects the object quantity specified in the text prompt, the cross-attention map at the early timesteps should form clusters of highly activated regions that exactly correspond to the intended object count, and each region should be clearly separated, as shown in \cref{figure1}(c).

In addition, existing methods \cite{binyamin2024count,kang2025counting,zafar2024iterativeobjectcountoptimization,zhang2023zeroshot} rely on external tools such as object counting and layout generation, or require training quantity-representations to encode instance count information.
As a result, they incur substantial computational costs and time overhead, making practical deployment of quantity control techniques challenging. Therefore, it is necessary to reflect quantity information without requiring additional modules and training.

In this paper, we propose \textit{CountCluster}, a method that generates the intended number of object instances without relying on any external tools or additional training. 
We found that clustering the object cross-attention map (CAM) at the first denoising timestep effectively controls the number of generated objects to match the intended count. We also observed that the clustering process should be guided by the originally high-activation regions in the CAM to reflect the intended object count.
Based on these observations, our method updates the latent representation at the first denoising timestep, guiding the object CAM to form spatially separated activation clusters that match the intended number of objects.
To achieve this, we construct a CAM-aware target object map by extracting and clustering the high-activation regions from the original CAM, ensuring that each cluster corresponds to a single intended object.
We then apply a Gaussian filter to the clustered map to obtain a continuous target distribution and compute the KL divergence between this distribution and the actual object CAM.
During the first denoising timestep, we minimize this KL divergence loss to guide the CAM toward spatially distinct activation regions that reflect the specified object count.

To demonstrate the effectiveness of our method, we conduct experiments across diverse prompt cases, validating its robustness and broad applicability.
The method consistently improves instance count accuracy over existing approaches.
Our approach achieves strong performance by ensuring the object CAM at the initial denoising timestep is well separated according to the intended object count, without requiring any additional training or external modules, ensuring both efficiency and ease of application.

\begin{figure*}[t]
\centering
\includegraphics[width=1.0\linewidth]{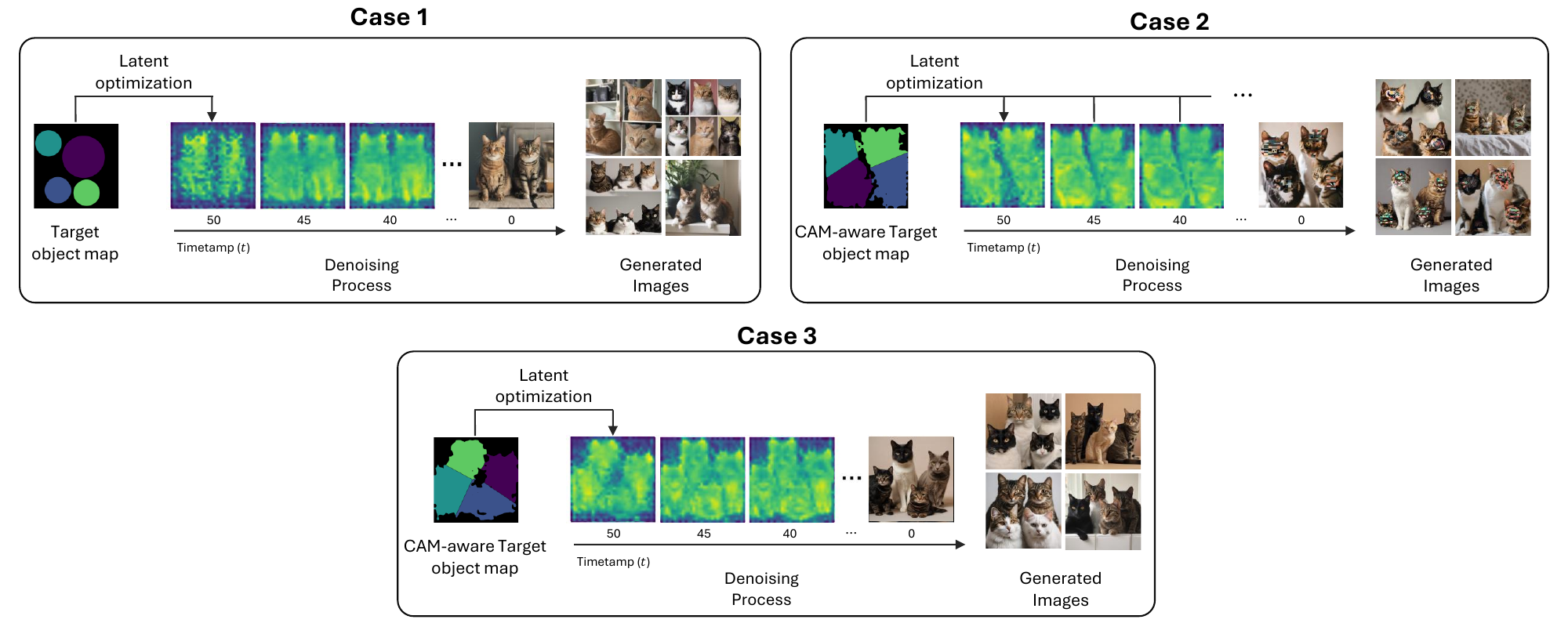}
\caption{Comparison of three latent optimization settings for controlling object quantity.}
\label{fig_motivation}
\vspace{-0.2cm}
\end{figure*}

\section{Related Work}
\subsection{Image Generation with Accurate Object Count}
Although diffusion-based text-to-image models have achieved remarkable progress in visual quality and semantic alignment, they still struggle to accurately reflect the number of objects explicitly specified in prompts \cite{Paiss_2023_ICCV,Hu_2023_ICCV}.
To address this limitation, various approaches have been proposed.
Iterative refinement methods \cite{binyamin2024count,kang2025counting} estimate the number of object instances during the image generation process, and then iteratively refine or regenerate the image based on the estimated count.
Another line of research focuses on explicitly incorporating quantity information into the generation process~\cite{zafar2024iterativeobjectcountoptimization,zhang2023zeroshot}.
These methods either introduce count tokens to directly train models for quantity control or adopt quantity-representation guided approaches, which extract numerical embeddings from the CLIP \cite{pmlr-v139-radford21a} feature space to encode object count information.
Nevertheless, these approaches still struggle with accurate object counting and even require additional training or external modules.

\subsection{Latent Optimization via Attention Guidance}
Training-free latent optimization has been actively explored to precisely control the outputs of pretrained text-to-image generation models.
In particular, various approaches leveraging cross-attention maps have been proposed.
Methods leveraging cross-attention maps are grounded in a simple observation: regions with high attention scores for a given text token tend to represent its concept more strongly in the generated image \cite{hertz2023prompttoprompt}.
Building on this insight, Attend-and-Excite \cite{10.1145/3592116} addresses the problem of subject neglect—where objects mentioned in the text prompt fail to appear in the generated image.
It introduces a loss function based on normalized cross-attention scores, which is used to iteratively optimize the latent representation.
Subsequently, several studies \cite{Agarwal_2023_ICCV,NEURIPS2023_0b08d733,Meral_2024_CVPR,Guo_2024_CVPR,2025selfcross} have proposed similar latent optimization strategies, primarily aiming to mitigate subject neglect by reinforcing attention to specific tokens.
Such methods have the advantage of regulating object presence without modifying the pretrained diffusion model.
However, they primarily focus on the binary presence or absence of objects, and still struggle to control quantitative attributes such as object count.
\section{Proposed Method}
\subsection{Preliminaries}
\label{sec_3_1}

\paragraph{Overview.}
We propose CountCluster, a method that generates the intended number of object instances without relying on any external tools or additional training. Our method updates the latent representation at the first denoising timestep, guiding the object CAM to form spatially separated activation clusters that match the intended number of objects.
We first introduce the background of attention-guided latent optimization in \cref{sec_3_1}.
Then, we present the motivations for generating objects that accurately match the intended count in \cref{sec_3_2} and introduce our proposed method in \cref{sec_3_3}.

\paragraph{Background of Attention-Guided Latent Optimization.}
Recent studies have revealed a strong correlation between the object cross-attention map (CAM) and the presence of specific objects in generated images.
Hertz et al.~\cite{hertz2023prompttoprompt} observed that, for a text token $y_i$, higher attention regions in its object CAM $\mathbf{A}_c^{(y_i)}$ tend to more strongly represent the corresponding visual concept in the generated image.
Building on this finding, several studies have proposed modifying CAMs to promote the emergence of desired objects and optimizing the latent representation accordingly~\cite{Agarwal_2023_ICCV,NEURIPS2023_0b08d733,Guo_2024_CVPR}.
These approaches design a loss function based on the object CAM to ensure that the object appears in the generated image, and iteratively update the latent variable $\mathbf{z}_t$ to minimize the loss.
To encourage the appearance of objects in the generated image, they generally perform latent optimization over multiple timesteps, typically during the early-to-middle stages of the denoising process.
This process can be formulated as follows:
\begin{equation}
\mathbf{z}'_t \leftarrow \mathbf{z}_t - \alpha_t \cdot \nabla_{\mathbf{z}_t} \mathcal{L}(\mathbf{A}_c^{(y_i)})
\label{eq:latent_update}
\end{equation}
where $\mathbf{z}_t$ denotes the latent representation at timestep $t$, and $\alpha_t$ is a scaling factor that controls the magnitude of the gradient step.

\subsection{Motivation}
\label{sec_3_2}
Recent research has shown that regions with higher activation in the object cross-attention map (CAM) are strongly correlated with the emergence of corresponding objects in generated images.
Building on this observation, we analyze how the object CAM can be effectively leveraged to control object quantities during the denoising process.

\Cref{fig_motivation} shows the results when the KL divergence between the object CAM and the target object map, which is clustered according to the intended number of objects, is used as a loss function for latent optimization.
When this KL divergence is minimized, the model tends to generate images that more strongly represent the corresponding visual concepts.
We conduct experiments under three different cases.
In case 1, randomly distributed clusters are used as the target object map, and latent optimization is applied only at the initial denoising timestep ($t=50$).
In case 2, a CAM-aware target object map is used, and latent optimization is performed across multiple timesteps (e.g., from 50 to 25).
In case 3, we apply CAM-aware latent optimization only once at the initial denoising step.

In case 1, randomly distributed clusters fail to reflect the intended object count. Because the random layout provides no meaningful spatial prior, the optimization gradients in latent space are misaligned with the diffusion model’s natural denoising process.
In contrast, case 2, which uses CAM-aware target object maps, successfully reflects the intended object count, showing that CAM-guided latent optimization can achieve quantity control without additional training.
However, in this setting, most generated objects tend to degrade image quality. This occurs because the appearance of objects in diffusion models is primarily determined in the very early timesteps, and repeatedly applying optimization in later timesteps interferes with this natural formation process.
In case 3, where latent optimization is performed only at the first denoising timestep, the generated images accurately reflect the intended count while maintaining natural structure and high visual quality.
Optimizing only at the initial timestep focuses on aligning the core semantic regions, thereby preserving both object count accuracy and overall image quality.

These experimental results lead to two key observations.
First, optimizing the latent representation such that the object CAM forms distinct clusters at the initial denoising timestep provides the most effective control over object quantity.
Second, the clustering process should be guided to form around the originally high-activation regions in the object CAM, ensuring that the semantic focus of each object is preserved while maintaining alignment with the intended count.
Based on these observations, we optimize the latent representation at the first denoising timestep, guiding the object CAM to form clearly separated activation clusters that match the intended number of objects.

\subsection{CountCluster}
\label{sec_3_3}
Based on the above observations, we propose \textit{CountCluster}, a method that generates the intended number of object instances without relying on any external tools or additional training.
During the first denoising timestep, we optimize the latent representation to minimize the KL divergence between the object CAM and a CAM-aware target object map, encouraging the model to produce spatially distinct activations that align with the intended object count.

To effectively reflect the intended count, the target object map should cluster the high-activation regions into the specified number of groups.
To achieve this, we utilize the high-activation regions from the object CAM.
We first extract regions whose activation scores exceed a predefined threshold and then cluster them to construct the CAM-aware target object map.
However, this target map is still binary, whereas cross-attention maps in diffusion models exhibit continuous-valued activations. 
To address this, we apply a Gaussian filter to the target map to obtain a continuous-valued target distribution.
Based on this extracted CAM-aware target object map, we compute the CountCluster loss to guide the object CAM to produce clearly separated activation clusters.
This approach enables the model to reflect the intended count while preserving high image quality, even in a fully training-free manner and without any external modules.

\subsubsection{CAM-aware Target Object Map Extraction}

\paragraph{High-Activation Region Extraction and Clustering.}
To precisely control the number of generated objects, it is essential that the clustering process is guided to form around the originally high-activation regions. To achieve this, we construct a CAM-aware target object map. We begin by restricting control to regions in the object CAM that exhibit sufficiently high-activation values. Specifically, we apply an attention threshold $\tau$ to the normalized CAM (we set $\tau = 0.35$) to generate a mask and extract candidate regions that are likely to contribute to object formation. These high-activation regions serve as the basis for determining the cluster centers, and our method guides object generation around these regions.

Our target map aims to obtain $k$ separated regions corresponding to the $k$ intended objects. To this end, we employ a connected component labeling algorithm to count the number of contiguous subregions and adjust this number to match the target object count $k$. If the number of regions exceeds $k$, we sequentially remove smaller regions. Conversely, if the number of regions is insufficient, we subdivide the largest region using a K-means–based partitioning and exclude the boundary between the two clusters from the mask, resulting in two independent regions. By iteratively applying this process, we obtain $k$ control regions that correspond to the intended object count while preserving the overall structure of the original CAM. Additional implementation details are provided in the Appendix.

\paragraph{Continuous CAM-aware Target Map Construction.}
In the previous step, we extracted $k$ high-activation cluster regions from the object CAM. However, the resulting target map is still a binary mask, which differs from the continuous activation patterns observed in diffusion models. To obtain a continuous target distribution, we apply a Gaussian filter to the clustered object map.

For each cluster, we construct a 2D Gaussian whose value is 1 at the cluster center and $\tau$ at the cluster boundary. The variance $\sigma$ is determined by the distance $r$ between the center and boundary:
\begin{equation}
\sigma = \sqrt{ \frac{r^2}{ -2 \log(\tau) } },
\qquad
T(\mathbf{x}) = \exp\left( -\frac{|\mathbf{x}-\boldsymbol{\mu}|^2}{2\sigma^2} \right).
\end{equation}
Here, $T(\mathbf{x})$ denotes the continuous CAM-aware target object map, $\boldsymbol{\mu}$ represents the cluster center, and $\mathbf{x}$ is the coordinate of each patch in the attention map.
This Gaussian smoothing produces coherent activation regions with clear separation between clusters, enabling the activation map to better reflect the intended object count.

\subsubsection{Latent Optimization with CountCluster Loss}
To guide the object CAM toward forming clearly separated activation clusters, we introduce the CountCluster loss, which consists of two components: an appearance loss $\mathcal{L}_{\mathrm{appear}}$ and a separation loss $\mathcal{L}_{\mathrm{sep}}$. 
The appearance loss encourages the object CAM to align with the CAM-aware target map composed of $k$ well-separated clusters, each corresponding to one intended object. 
The separation loss suppresses activation outside the cluster regions, maintaining clear boundaries between objects and keeping the object CAM focused on the intended object areas.

To align the object CAM with the target map, we measure the discrepancy between the actual CAM $A_c(x)$ and the target object map $T_j(x)$ using KL divergence~\cite{kullback1951information}.  
The appearance loss is formulated as:
\begin{equation}
\mathcal{L}_{\mathrm{appear}} = \frac{1}{\sqrt{k}} \sum_{j=1}^k D_{\mathrm{KL}}(T_j(x) \,\|\, A_c(x))
\label{eq:kl}
\end{equation}
where the KL divergence is computed independently for each of the $k$ clusters.
Since the total KL value naturally increases with the number of clusters $k$, we normalize it by $\sqrt{k}$ to maintain a consistent scale across different object counts.
To suppress unnecessary activation outside the clusters, we restrict attention values in non-target regions so that they do not exceed the threshold $\tau$.
Since a direct max operation is non-differentiable, we employ a smooth approximation using the LogSumExp function:
\begin{equation}
\mathcal{L}_{\mathrm{sep}} = \mathrm{ReLU}\!\left(\mathrm{LogSumExp}(\mathbf{A}_c^{(y_i)} - \tau)\right)
\label{eq:L_out}
\end{equation}
This formulation suppresses overly high-activation outside the cluster mask and guides the CAM to remain focused within object-relevant regions.
Finally, the total loss is consisted of the appearance and separation losses, and applied at the first denoising timestep to update the latent representation:
\begin{equation}
\mathcal{L}_{\mathrm{total}} = \mathcal{L}_{\mathrm{appear}} + \mathcal{L}_{\mathrm{sep}},
\end{equation}
\begin{equation}
\mathbf{z}'_t \leftarrow \mathbf{z}_t - \alpha_t \cdot \nabla_{\mathbf{z}_t} \mathcal{L}_{total}
\label{eq:latent_update}
\end{equation}
\section{Experiments}
\subsection{Experimental setup}

\paragraph{Benchmarks.}
To quantitatively evaluate the model’s ability to control object quantities, we utilized various quantity-aware prompt sets following previous object-counting studies~\cite{binyamin2024count,kang2025counting}. Each prompt follows the template \textit{a photo of [count] [object]''}, where \textit{[count]} is a number word ranging from \textit{two''} to \textit{``ten''}. This design allows systematic evaluation across diverse object categories and a range of object counts.
To demonstrate the method’s robustness and adaptability across diverse prompt conditions, we also evaluate the proposed method on more complex scenarios, including coiled and elongated objects, occluded scenes, and various natural backgrounds, as discussed in \cref{sec_5_2}.

\begin{figure*}[t]
\centering
\includegraphics[width=1\linewidth]{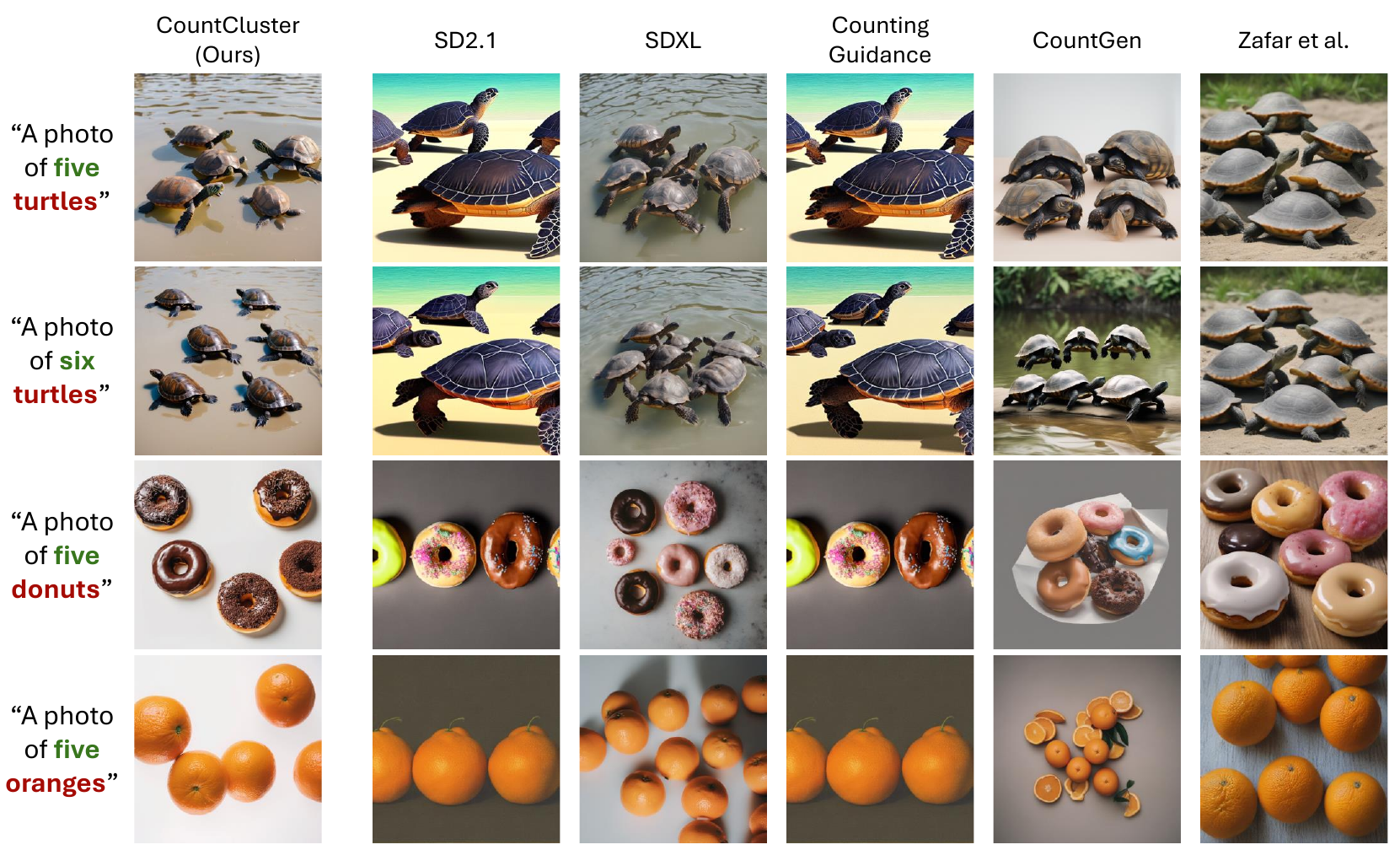}
\caption{
The first two rows illustrate cases where only the object quantity in the prompt is changed while keeping the random seed.
The last two rows show results where only the object category (e.g., “donuts” → “oranges”) is changed under the same random seed.
}
\label{fig_main}
\end{figure*}

\paragraph{Baselines.}
We compare our method against five baseline approaches, including those specifically designed for object quantity control.
The baselines are implemented on either SD2.1~\cite{Rombach_2022_CVPR} or SDXL~\cite{podell2024sdxl}, and results are reported separately according to the underlying diffusion backbone.
Counting-Guidance~\cite{kang2025counting} operates on SD2.1, whereas Zafar et al.~\cite{zafar2024iterativeobjectcountoptimization} and CountGen~\cite{binyamin2024count} are built on SDXL.
Since our method requires no architectural changes, training, or auxiliary modules, it can be directly applied to both backbones.
For fair comparison, we therefore evaluate and report our method on both SD2.1 and SDXL alongside their corresponding baselines.

\paragraph{Metrics.}
For evaluation, we assess the proposed method from two complementary perspectives: object-count accuracy and image quality.
For counting evaluation, we employ two independent assessment tools—CountGD~\cite{10.5555/3737916.3739463}, an object-counting model built upon Grounding DINO~\cite{10.1007/978-3-031-72970-6_3}, and the TIFA framework~\cite{Hu_2023_ICCV}, which uses GPT-4o mini~\cite{openai2024gpt4ocard} as its Visual Question Answering (VQA) backbone.
Both tools automatically estimate the number of object instances in each generated image, and the results are compared with the quantities specified in the prompts to measure quantitative accuracy.
For image-quality evaluation, we adopt CLIPScore~\cite{hessel2022clipscorereferencefreeevaluationmetric} and ImageReward~\cite{xu2023imagerewardlearningevaluatinghuman} to assess visual-textual quality and human-preference alignment.

\begin{table*}[t]
\centering
\footnotesize
\begin{tabular}{l|ccc|cc}
\toprule 
\multirow{2}{*}{Method} & \multicolumn{3}{c|}{CountGD~\cite{10.5555/3737916.3739463}} & \multicolumn{2}{c}{TIFA~\cite{Hu_2023_ICCV}} \\ \cline{2-6} \rule{0pt}{1.2EM}
& Acc. ($\uparrow$) & MAE ($\downarrow$) & RMSE ($\downarrow$) & Counting ($\uparrow$) & Appearing ($\uparrow$)
\\ \midrule
\multicolumn{6}{c}{SD2.1~\cite{Rombach_2022_CVPR}}
\\ \midrule
Naive SD2.1~\cite{Rombach_2022_CVPR} & 24.56 & 1.990 & 3.127 & 72.15 & 99.75 \\
Counting-Guidance~\cite{kang2025counting} & 24.95 & 1.885 & \textbf{2.982} & 71.57 & 99.51\\
Ours & \textbf{33.33} & \textbf{1.791} & 3.282 & \textbf{75.76} & \textbf{99.26} \\ \midrule
\multicolumn{6}{c}{SDXL~\cite{podell2024sdxl}}
\\ \midrule
Naive SDXL~\cite{podell2024sdxl} & 27.88 & 5.045 & 9.302 & 73.01 & \textbf{99.75} \\
Zafar et al.~\cite{zafar2024iterativeobjectcountoptimization} & 25.93 & 1.815 & 2.924 & 76.62 & 99.01  \\
CountGen~\cite{binyamin2024count} & 46.20 & 1.661 & 3.562 & 76.48 & 97.28 \\
Ours & \textbf{55.75} & \textbf{0.821} & \textbf{1.849} & \textbf{79.51} & 99.26
\\ \bottomrule
\end{tabular}
\caption{Experimantal Results on SD2.1 and SDXL using CountGD \cite{10.5555/3737916.3739463} and TIFA \cite{Hu_2023_ICCV}.
CountGD measures numerical accuracy using Accuracy (↑), MAE (↓), and RMSE (↓),
while TIFA evaluates text–image consistency through Counting (↑) and Appearing (↑) scores.
}
\label{tab_1}
\end{table*}

\begin{table}[t]
\centering
\footnotesize
\begin{tabular}{l|cc}
\toprule 
\multirow{2}{*}{Method} & \multicolumn{2}{c}{Image Quality} \\ \cline{2-3} \rule{0pt}{1.2EM}
  & CLIPScore ($\uparrow$) & ImageReward ($\uparrow$)
\\ \midrule
\multicolumn{3}{c}{SD2.1~\cite{Rombach_2022_CVPR}}
\\ \midrule
Naive SD2.1~\cite{Rombach_2022_CVPR} & 0.265 & 0.780
\\
Counting Guidance~\cite{kang2025counting} & 0.265 & 0.768
\\
Ours & 0.265 & 0.744
\\ \midrule
\multicolumn{3}{c}{SDXL~\cite{podell2024sdxl}}
\\ \midrule
Naive SDXL~\cite{podell2024sdxl} & 0.268 & 0.568
\\
Zafar et al.~\cite{zafar2024iterativeobjectcountoptimization} & 0.267 & 0.921
\\
CountGen~\cite{binyamin2024count} & 0.268  & 0.550
\\
Ours & 0.262 & 0.563
\\  \bottomrule
\end{tabular}
\caption{Comparison of image quality using CLIPScore~\cite{hessel2022clipscorereferencefreeevaluationmetric} and ImageReward~\cite{xu2023imagerewardlearningevaluatinghuman}.}
\vspace{-0.1cm}
\label{tab_2}
\end{table}

\begin{table}[t]
\centering
\footnotesize
\setlength{\tabcolsep}{3pt}
\begin{tabular}{l|cccc}
\toprule
Method & Time (s) & Mem. (GB) & Training & External
\\ \midrule
\multicolumn{5}{c}{SD2.1~\cite{Rombach_2022_CVPR}}
\\ \midrule
Naive SD2.1~\cite{Rombach_2022_CVPR}& 5.89 &  7 & \xmark & \xmark
\\
Counting Guidance~\cite{kang2025counting} & 21.27 & 14 & \xmark & \cmark
\\
Ours & 9.14 & 7 & \xmark & \xmark
\\ \midrule
\multicolumn{5}{c}{SDXL~\cite{podell2024sdxl}}
\\ \midrule
Naive SDXL~\cite{podell2024sdxl} & 13.32 & 15 & \xmark & \xmark
\\
Zafar et al.~\cite{zafar2024iterativeobjectcountoptimization} & 9.07 & 16 (26$^{*}$) & \cmark & \xmark
\\
CountGen~\cite{binyamin2024count} & 88.04 & 71 & \xmark & \cmark
\\
Ours & 36.38 & 18 & \xmark & \xmark
\\  \bottomrule
\end{tabular}
\caption{Inference time and peak GPU memory are measured per generated image on an RTX 3090. “Training’’ indicates whether additional model fine-tuning is required, and “External’’ denotes reliance on auxiliary counting or detection modules.}
\label{tab_3}
\vspace{-0.2cm}
\end{table}

\subsection{Qualitative Results}
\Cref{fig_main} presents a qualitative comparison between our method and existing baselines.
To evaluate how models respond to changes in object count, we generated images from prompts that differ only in the specified number while keeping the initial latent fixed. Our method reliably reflects the intended quantity (e.g., five turtles → six turtles) without disrupting scene composition or visual coherence. In contrast, baseline models often fail to adjust the count or generate duplicated and overlapping objects.

Our method also maintains accurate instance counts across diverse object categories (e.g., five donuts → five oranges). These results demonstrate that our method achieves robust quantity control through attention-based clustering rather than relying solely on pretrained model priors.
Overall, the qualitative results show that our method preserves visual fidelity while producing images that adhere to the quantities specified in the prompt.

\subsection{Quantitative Results}
\paragraph{Object Count Accuracy.}
~\Cref{tab_1} presents the quantitative comparison on SD2.1 and SDXL using CountGD~\cite{10.5555/3737916.3739463} and TIFA~\cite{Hu_2023_ICCV}.
Across all metrics—including Accuracy, MAE, RMSE, and VQA-based counting consistency—our proposed \textit{CountCluster} outperforms existing approaches.
For SDXL, our method also substantially outperforms numerical precision over prior methods.
In addition, the TIFA results confirm that our method maintains strong text–image alignment.
These improvements indicate that aligning early-timestep attention clustering effectively enhances quantitative faithfulness between prompt and image without requiring extra training.

\paragraph{Image Quality.}
~\Cref{tab_2} reports the image quality evaluation using CLIPScore~\cite{hessel2022clipscorereferencefreeevaluationmetric} and ImageReward~\cite{xu2023imagerewardlearningevaluatinghuman}.
Although our approach focuses solely on quantity control, it maintains comparable visual quality to baseline models.
The results show that our method achieves nearly identical CLIPScore while exhibiting only minor variation in ImageReward, confirming that quantitative control does not compromise perceptual quality or prompt-image coherence.

\paragraph{Efficiency Analysis.}
~\Cref{tab_3} compares inference time, GPU memory, and dependency overhead.
Our method introduces only minimal computational cost compared to base diffusion models, while avoiding any additional training or external counting modules.
Unlike prior methods such as CountGen, which require up to 71 GB of memory and multi-stage regeneration, our method operates in a single forward denoising pass with negligible overhead.
Overall, our method achieves accurate and efficient object-count control while remaining fully training-free and independent of external modules.

\begin{table}[t]
\centering
\footnotesize
\begin{tabular}{l|ccc}
\toprule 
\multirow{2}{*}{Method} & \multicolumn{3}{c}{CountGD} \\ \cline{2-4} \rule{0pt}{1.2EM}
& Acc. ($\uparrow$) & MAE ($\downarrow$) & RMSE ($\downarrow$) \\ \midrule
Naive SDXL & 27.88 & 5.045 & 9.302 \\ \midrule
Case 1 & 29.82 & 3.353 & 7.137 \\
Case 2 & 33.33 & 54.493 & 114.855 \\ \midrule
w/o $L_{appear}$ & 20.66 & 6.544 & 12.503 \\
w/o $L_{sep}$ & \underline{42.88} & \underline{0.949} & \textbf{1.553} \\ \midrule
Ours & \textbf{55.75} & \textbf{0.821} & \underline{1.849} 
\\ \bottomrule
\end{tabular}
\caption{Ablation study on key components of CountCluster.}
\label{tab_abla}
\end{table}

\begin{figure}[t]
\centering
\includegraphics[width=0.8\linewidth]{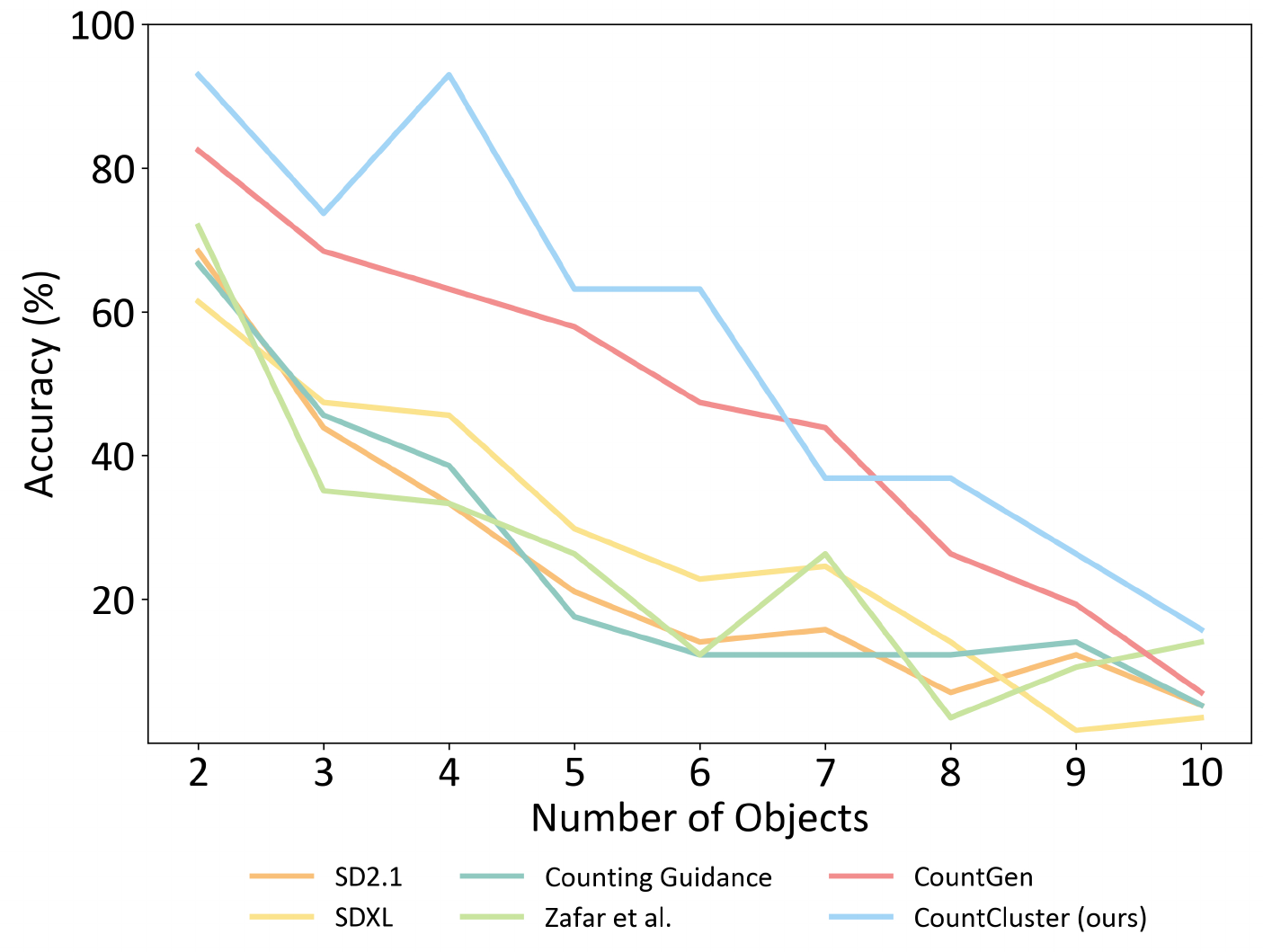}
\vspace{-0.3cm}
\caption{Accuracy comparison across different object counts. It illustrates the model performance as the number of objects specified in the prompt increases from 2 to 10.
}
\vspace{-0.3cm}
\label{figure4}
\end{figure}

\section{Analysis}
\subsection{Ablation Study}
Table~\ref{tab_abla} shows the contribution of each component in CountCluster. Cases 1 and 2 follow the same settings described in \cref{sec_3_2}. Using a random Gaussian target cluster map (Case 1) yields only modest improvement over the naive model, indicating that the target distribution must align with meaningful high-activation regions. Applying latent optimization across multiple early timesteps (Case 2) leads to highly unstable predictions and large RMSE values, confirming that optimization beyond the first denoising step disrupts the model's natural generation trajectory.

\begin{figure*}[t!]
\centering
\includegraphics[width=1.0\linewidth]{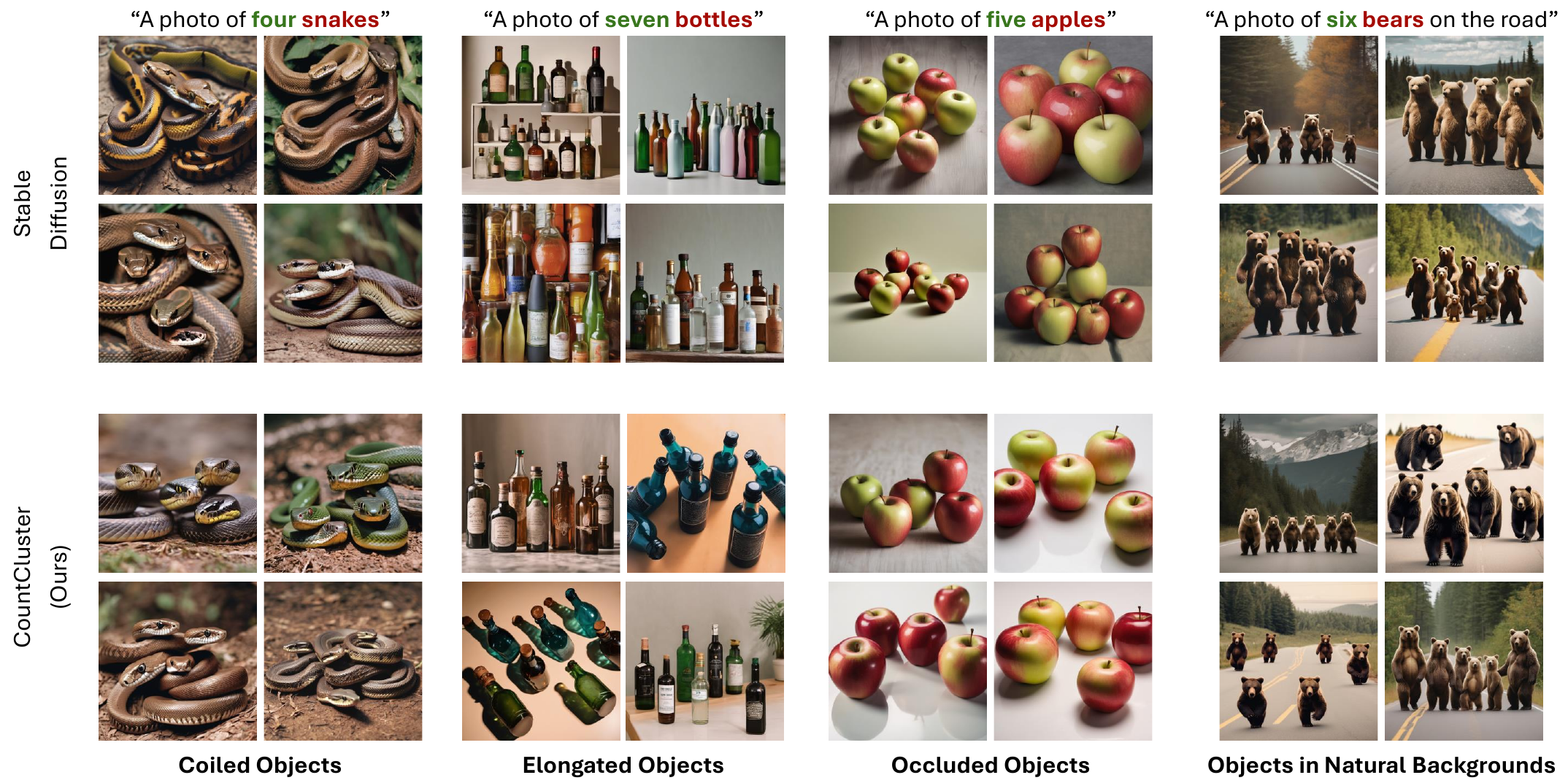}
\caption{Qualitative results on complex object configurations, including coiled, elongated, occluded, and objects in natural backgrounds.
}
\label{figure5}
\end{figure*}

We further examine the contribution of each loss component in the CountCluster loss.
Removing the appearance term prevents the object CAM from aligning with the CAM-aware target map, weakening cluster formation and causing the model to frequently underestimate the object count.
In contrast, removing the separation term allows activation to spread beyond the target regions, producing merged or duplicated attention clusters that lead to over-counting.
Using both terms together yields the most stable cluster structure and the best quantitative performance, indicating that appearance alignment and activation suppression are both essential for reliable quantity control.

\subsection{Discussions}
\label{sec_5_2}

\paragraph{Accuracy via Object Counts.}
\Cref{figure4} compares the counting accuracy of different methods across varying object counts.
All methods show reduced accuracy as the number of objects increases, reflecting the challenge of maintaining instance separation in crowded scenes.
Our method achieves consistently higher accuracy with a smoother degradation curve, matching CountGen at two objects and surpassing all baselines beyond three.
Notably, accuracies peak around four and nine objects, likely due to regular spatial layouts (e.g., 2$\times$2 or 3$\times$3 grids) that diffusion models naturally favor.
These results demonstrate that our method maintains quantitative consistency even under complex spatial configurations.

\paragraph{Qualitative Results on Complex Prompts.}
\Cref{figure5} presents qualitative results on complex prompts.
The proposed method maintains accurate quantity control under diverse spatial configurations and visual contexts, including coiled, elongated, occluded, and natural-scene objects.
Even when instances overlap or blend into the background, our method effectively separates attention regions, preventing merging or omission.
Since latent optimization is performed only at the first denoising timestep, objects are naturally distributed while preserving spatial consistency guided by the target cluster map.
These results confirm that our approach achieves visually coherent and robust quantity control across challenging scenarios.
\section{Conclusion}
In this work, we showed that object quantity in diffusion models is strongly influenced by the object cross-attention map (CAM) at the first denoising timestep. Leveraging this insight, we introduced CountCluster, a training-free approach that controls object count without external modules. By clustering high-activation CAM regions and constructing a continuous CAM-aware target map, the method guides early-timestep CAMs toward activation patterns that match the intended object count through a KL-based optimization.
CountCluster requires no training or architectural changes yet effectively aligns generated images with the specified number of objects while maintaining visual quality across diverse prompt settings.

{
    \small
    \bibliographystyle{ieeenat_fullname}
    \bibliography{main}
}

\clearpage
\twocolumn[
{
    \begin{center}
    {\Large \textbf{CountCluster: Training-Free Object Quantity Guidance with Cross-Attention \\[0.25em] Map Clustering for Text-to-Image Generation}}\\[0.75em]
    \Large Supplementary Material

    \vspace{1.5em}
    \refstepcounter{figure}\label{figure6}
    \includegraphics[width=0.85\linewidth]{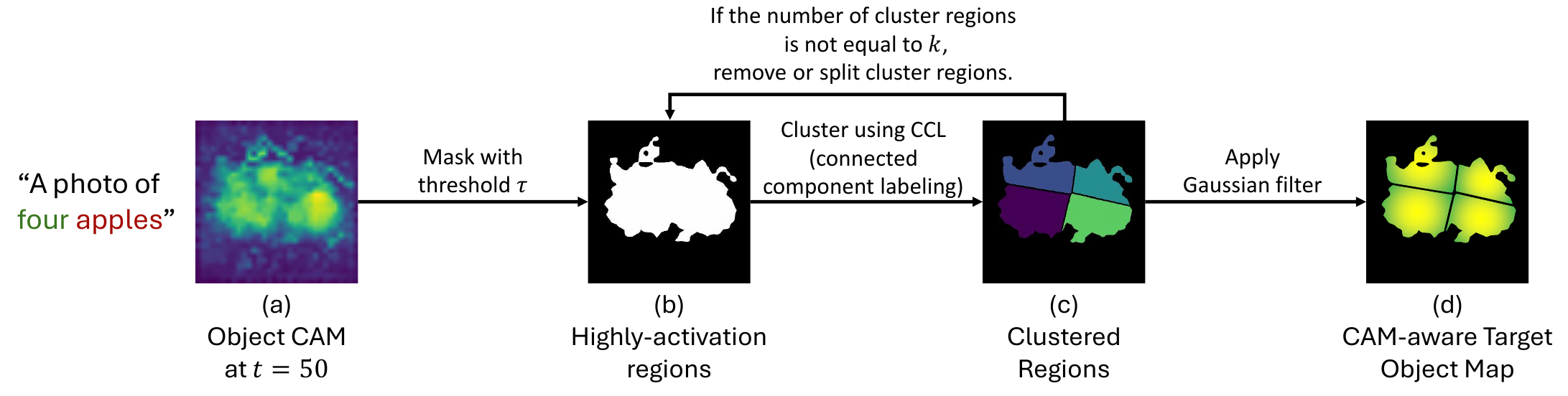}
    \vspace{0.5em}

    \begin{flushleft}
    {\small Figure 6.
    Overview of the CAM-aware target object map extraction process, including CAM masking, cluster formation via CCL, cluster \\[-0.75em] count adjustment, and Gaussian filtering to obtain the target object map.}
    \end{flushleft}

    \vspace{1em}
    \end{center}
}
]
\setcounter{section}{0}
\renewcommand{\thesection}{\Alph{section}}

\section{Details of the Proposed Method}
In this section, we explain the details of the proposed method.
As described in the Motivation section, to generate the intended number of object instances without additional training, the object cross-attention map (CAM) needs to form distinct clusters around its high-activation regions at the initial denoising timestep.
Therefore, we extract a CAM-aware target object map in which the object CAM forms clear and well-separated clusters for the intended number of objects, and then optimize the latent representation by minimizing the KL divergence between the object CAM and this CAM-aware target object map.

In CAM-aware target object map extraction, we first apply Connected Component Labeling (CCL) to the object CAM to divide it into separate regions. If the number of resulting regions exceeds the intended object count, we remove smaller regions to match the target count. In contrast, if the number of regions is smaller than intended, we split larger regions to obtain the required number. The resulting regions are converted into a continuous distribution by applying a Gaussian filter. Finally, based on the CAM-aware target object map, we optimize the latent representation using the appearance loss and the separation loss.

\begin{figure}[t]
    \centering
    \subfloat[\centering{Case with fewer clusters than the intended count}]
    {\includegraphics[width=1.0\linewidth]{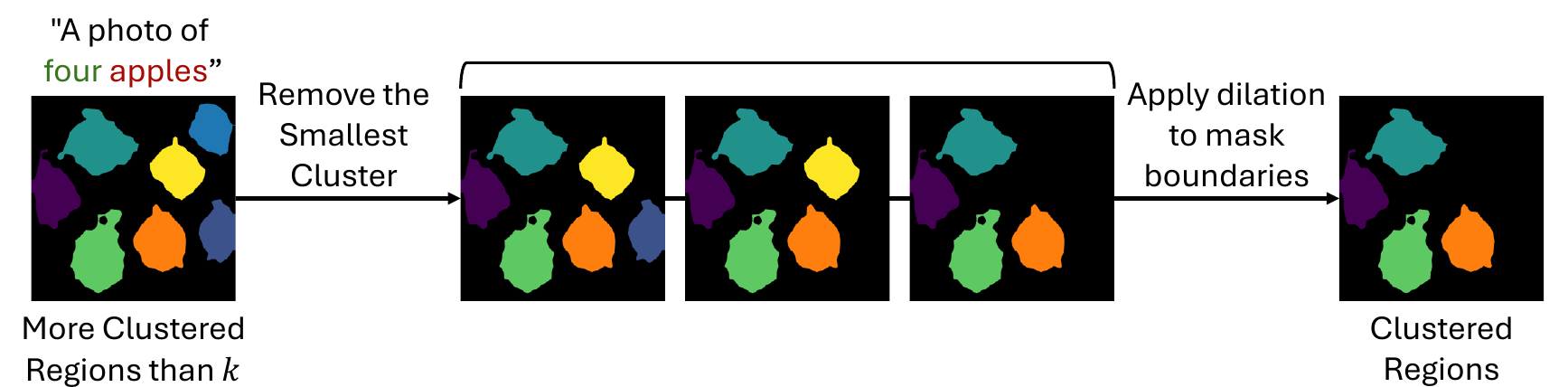}\label{fig7_a}}
    \vspace{0.8em}

    \subfloat[\centering{Case with more clusters than the intended count}]
    {\includegraphics[width=1.0\linewidth]{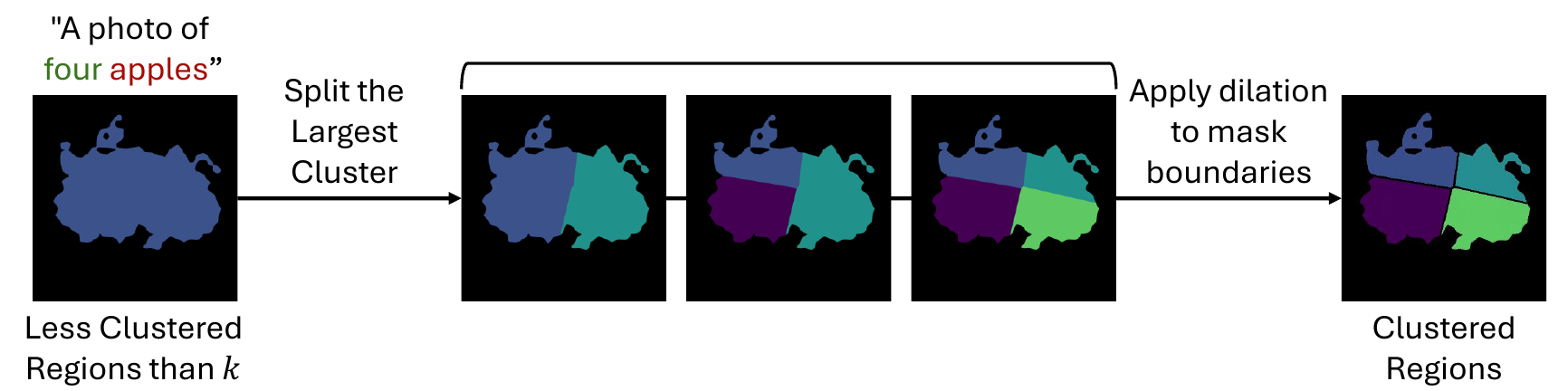}\label{fig7_b}}
    
    \caption{Cluster count adjustment process to match the intended count. 
    (a) Case with fewer clusters than the intended count. 
    (b) Case with more clusters than the intended count.}
    \label{figure7}
\end{figure}

\subsection{CAM-aware Target Object Map Extraction}
\cref{figure6} illustrates the overall process of CAM-aware target object map extraction.
To construct a target map where clusters form around the originally high-activation regions, we first mask out the regions of the object CAM with activation values lower than the threshold $\tau$, so that only regions above $\tau$ remain, as shown in \cref{figure6} (b).

Next, we apply connected component labeling (CCL) to form the intended $k$ separated object CAM clusters. 
CCL is a conventional image-processing technique that assigns distinct labels to separated activation regions based on DFS searching, enabling the extraction of separated activation clusters.
However, CCL does not guarantee that the number of detected clusters matches the intended count $k$.
Therefore, we refine the clustering output to align with the intended number of clusters, as shown in \cref{figure6} (c).
If the number of resulting regions exceeds $k$, we remove smaller clusters until the target count is reached, as illustrated in \cref{figure7} (a). In contrast, if the number of regions is smaller than $k$, we split larger clusters using a k-means–based partitioning process until the target count is reached, as shown in \cref{figure7} (b). Through this iterative process, we obtain $k$ well-separated cluster regions.

Based on $k$-clustered regions, we apply a Gaussian filter to the target map to obtain a continuous-valued target distribution. This CAM-aware target object map guides the latent optimization process for object count control.

\subsection{Latent Optimization with CountCluster Loss}
Based on the extracted CAM-aware target object map, we compute the CountCluster loss, which consists of appearance and separation losses, and optimize the latent representation with our loss.

The appearance loss is defined as the KL divergence~\cite{kullback1951information} between the object CAM \(A_c(x)\) and the target object map \(T_j(x)\).
Minimizing this loss encourages the object CAM to become closer to the target map, allowing it to form $k$ clearly separated activation clusters.

The separation loss is computed as the softmax value of non–target-map regions, regions where no cluster is located.
When the separation loss is minimized, the model suppresses CAM activations in non–target-map regions below the threshold $\tau$, which reduces unnecessary activation outside the clusters and leads to clearer separation between clusters. Using our proposed loss, we update the latent representation at the first denoising timestep.

\section{Implementation Details}
\subsection{Detailed Experimental Setup}
In this study, we implemented our method and determined all hyperparameters with reference to the diffusion pipeline code of Attend-and-Excite \cite{10.1145/3592116}. The proposed quantity control method was build on both pretrained SD2.1 \cite{Rombach_2022_CVPR} and SDXL \cite{podell2024sdxl} models. The object tokens and their quantities were automatically extracted from the prompt using the \texttt{inflect} library.

During image generation, the guidance scale \cite{ho2022classifierfreediffusionguidance} of the diffusion model was fixed at 7.5. In the proposed method, Gaussian smoothing with a kernel size of $3$ and $\sigma=0.5$ was applied to the object cross-attention map, followed by min-max normalization. For the clustering step, we set the attention score threshold $\tau=0.35$.

The image latent was updated during the first 10 timesteps of the denoising process ($t=50$). In ours (SD2.1), the scale factor ($\alpha$) was set to 1, while in ours (SDXL), it was set to 10,000. The latent was updated 20 times in total.

For image generation, CountGen \cite{binyamin2024count} utilized four NVIDIA RTX 3090 GPUs in parallel for inference, whereas all other experiments were conducted on a single RTX 3090 GPU. In addition, count token training for the method of Zafar et al. \cite{zafar2024iterativeobjectcountoptimization} was performed using two RTX 3090 GPUs.

\subsection{Detailed Benchmark Description}
For our experiments, we constructed a benchmark prompt set by integrating those proposed in Counting Guidance \cite{kang2025counting} and CountGen \cite{binyamin2024count}, using prompts of the form \textit{``A photo of [count] [object]''}. Here, \texttt{[count]} is an English word from two to ten, and \texttt{[object]} is selected from the 19 predefined categories listed in \cref{tab:promptset}.

\begin{table}[h]
\centering
\small
\begin{tabular}{p{0.11\linewidth} | p{0.65\linewidth}}
\toprule
\textbf{Count} &
\texttt{two}, \texttt{three}, \texttt{four}, \texttt{five}, \texttt{six}, \texttt{seven}, \texttt{eight}, \texttt{nine}, \texttt{ten} \\
\midrule
\textbf{Object} &
\texttt{apples}, \texttt{bears}, \texttt{birds}, \texttt{cars}, \texttt{cats}, \texttt{clocks}, \texttt{dogs}, \texttt{donuts}, \texttt{eggs}, \texttt{elephants}, \texttt{frogs}, \texttt{horses}, \texttt{lemons}, \texttt{mice}, \texttt{monkeys}, \texttt{onions}, \texttt{oranges}, \texttt{tomatoes}, \texttt{turtles} \\
\bottomrule
\end{tabular}
\caption{List of counts and objects used in the benchmark set}
\label{tab:promptset}
\end{table}

\subsection{Evaluation Metrics}
\paragraph{Counting Module-based Evaluation.}
To evaluate the quantity of objects in the generated images, we employed CountGD \cite{10.5555/3737916.3739463} to automatically measure the number of instances in each image corresponding to the object specified in the prompt. The predicted number of instances obtained from CountGD were then compared with the target counts provided in the prompts to compute the following evaluation metrics: Accuracy (Acc), Mean Absolute Error (MAE), and Root Mean Square Error (RMSE). The formulas for each metric are as follows:
\begin{equation}
\text{Acc} = \sum_{i=1}^{n} \frac{ \mathbf{1}[y_i = \hat{y}_i] }{n}
\end{equation}

\begin{equation}
\text{MAE} = \sum_{i=1}^{n} \frac{ |y_i - \hat{y}_i| }{n}
\end{equation}

\begin{equation}
\text{RMSE} = \sqrt{ \sum_{i=1}^{n} \frac{ (y_i - \hat{y}_i)^2 }{n} }
\end{equation}
Here, $y_i$ denotes the target object count specified in the $i$-th prompt, and $\hat{y}_i$ is the number of corresponding object instances in the $i$-th generated image as predicted by CountGD. $n$ represents the total number of generated images. In our experiments, we evaluated each method on a total of 1,710 images, generated by applying 10 random seeds (0$\sim$9) to all 171 combinations of counts and objects.

\paragraph{VQA-Based Evaluation.}
For VQA-based evaluation, we utilize the TIFA framework \cite{Hu_2023_ICCV}, which utilizes GPT-4o mini \cite{openai2024gpt4ocard} as the VQA model, to jointly evaluate the Counting score and Appearing score. The TIFA score is computed as the proportion of VQA queries for which GPT-4o mini provides the correct answer based on the generated image:
\begin{equation}
\text{TIFA Score} = \sum_{i=1}^{n} \frac{ \mathbf{1}[A_i = \hat{A}_i] }{n}
\end{equation}
Here, $A_i$ denotes the ground-truth answer for the $i$-th question, $\hat{A}_i$ represents the answer generated by GPT-4o mini, and $n$ is the total number of question-answer pairs.
The question-answer pairs actually used in the TIFA framework are exemplified in \cref{tab:Counting_VQA} and \cref{tab:Appearing_VQA}.

\begin{table}[h]
\centering
\small
\begin{tabular}{p{0.9\linewidth}}
\toprule
\textbf{Prompt}: \textit{A photo of two apples} \\[2pt]
\hline
\noalign{\vskip 2pt}
\textbf{Question}: How many apples are in the photo? \\[2pt]
\textbf{Choices}: 1, 2, 3, 4 \\[2pt]
\hline
\noalign{\vskip 2pt}
\textbf{Answer}: 2 \\
\bottomrule
\end{tabular}
\caption{Example of Counting VQA}
\label{tab:Counting_VQA}
\end{table}

\begin{table}[h]
\centering
\small
\begin{tabular}{p{0.9\linewidth}}
\toprule
\textbf{Prompt}: \textit{A photo of two apples} \\[2pt]
\hline
\noalign{\vskip 2pt}
\textbf{Question}: What fruit is in the photo? \\[2pt]
\textbf{Choices}: apples, oranges, bananas, pears \\[2pt]
\hline
\noalign{\vskip 2pt}
\textbf{Answer}: apples \\
\bottomrule
\end{tabular}
\caption{Example of Appearing VQA}
\label{tab:Appearing_VQA}
\end{table}

\paragraph{Image Quality Evaluation.} 
To evaluate the quality of the generated images, we employ ImageReward \cite{xu2023imagerewardlearningevaluatinghuman} and CLIPScore \cite{hessel-etal-2021-clipscore}.
ImageReward captures human preference for image quality. ImageReward is a reward model trained on expert preference comparisons over roughly 130,000 text–image pairs. It considers multiple aspects such as image quality, text–image alignment, and aesthetics to approximate human preference. For each prompt–image pair, the model outputs a scalar score, and higher scores mean that the image is more preferred by humans.
In contrast, CLIPScore measures how well the image semantically aligns with the text. It is computed using the cosine similarity between the text embedding 
$c$ and image embedding $v$ obtained from CLIP \cite{pmlr-v139-radford21a}. CLIPScore is defined as:
\begin{equation}
\text{CLIPScore}(c, v) = w \cdot \max\left( \cos(c, v), 0 \right),
\end{equation}
where $w=2.5$ is a scaling factor used to normalize the score to the $[0, 1]$ range. Higher CLIPScore values indicate stronger semantic alignment between the generated image and the textual prompt.

\begin{table}[t]
\centering
\footnotesize
\begin{tabular}{l|ccc}
\toprule 
\multirow{2}{*}{Method} & \multicolumn{3}{c}{CountGD} \\ \cline{2-4} \rule{0pt}{1.2EM}
& Acc. ($\uparrow$) & MAE ($\downarrow$) & RMSE ($\downarrow$) \\ \midrule
GMM~\cite{Reynolds2009} & 47.56 & 0.953 & \textbf{1.674} \\
Ours & \textbf{55.75} & \textbf{0.821} & 1.849
\\ \bottomrule
\end{tabular}

\caption{Ablation study of GMM and CCL (ours) based target map extraction on counting accuracy.}
\label{tab_GMM1}
\end{table}

\begin{table}[t]
\centering
\footnotesize
\begin{tabular}{l|cc}
\toprule 
\multirow{2}{*}{Method} & \multicolumn{2}{c}{Image Quality} \\ \cline{2-3} \rule{0pt}{1.2EM}
  & CLIPScore ($\uparrow$) & ImageReward ($\uparrow$)
\\ \midrule
GMM & \textbf{0.269} & 0.211 \\
Ours & 0.262 & \textbf{0.563} \\  \bottomrule
\end{tabular}
\caption{Ablation study of GMM and CCL (ours) based target map extraction on image quality.}
\vspace{-0.1cm}
\label{tab_GMM2}
\end{table}

\begin{figure}[ht]
    \centering
    
    \subfloat[\centering{Objects that are closely placed or occluded}]
    {\includegraphics[width=1.0\linewidth]{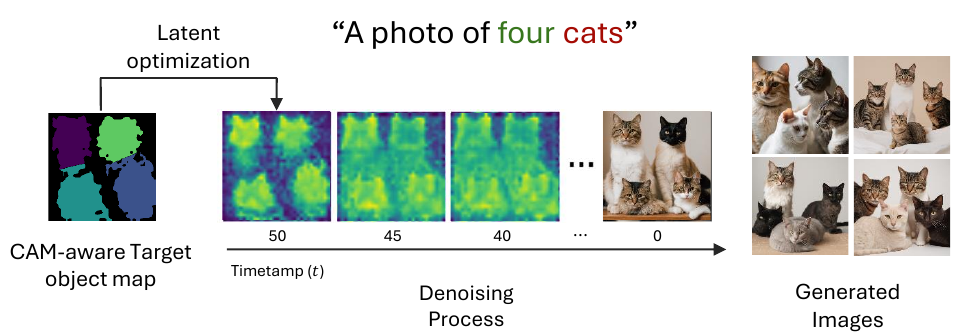}\label{fig8_a}}
    
    \vspace{0.8em}
    
    \subfloat[\centering{Objects with non-Gaussian shapes}]
    {\includegraphics[width=1.0\linewidth]{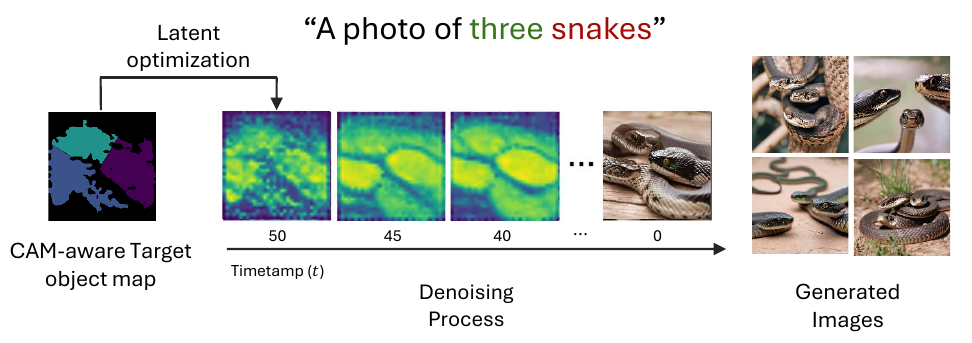}\label{fig8_b}}
    
    \caption{Images generated by our method under challenging scenarios.
    (a) Objects that are closely placed or occluded.
    (b) Objects with non-Gaussian shapes.}
    \label{figure8}
\end{figure}

\begin{table*}[t]
\centering
\footnotesize
\begin{tabular}{l|ccc|cc}
\toprule 
\multirow{2}{*}{Method} & \multicolumn{3}{c|}{CountGD~\cite{10.5555/3737916.3739463}} & \multicolumn{2}{c}{TIFA~\cite{Hu_2023_ICCV}} \\ \cline{2-6} \rule{0pt}{1.2EM}
& Acc. ($\uparrow$) & MAE ($\downarrow$) & RMSE ($\downarrow$) & Counting ($\uparrow$) & Appearing ($\uparrow$)
\\ \midrule
\multicolumn{6}{c}{SD2.1~\cite{Rombach_2022_CVPR}}
\\ \midrule
Naive SD2.1~\cite{Rombach_2022_CVPR} & 30.00 & \textbf{1.460} & \textbf{2.126} & 71.04 & \textbf{82.41} \\
Counting-Guidance~\cite{kang2025counting} & 35.00 & 2.000 & 6.198 & 73.40 & 82.41\\
Ours & \textbf{39.50} & 2.045 & 8.229 & \textbf{74.41} & 81.41 \\ \midrule
\multicolumn{6}{c}{SDXL~\cite{podell2024sdxl}}
\\ \midrule
Naive SDXL~\cite{podell2024sdxl} & 27.50 & 8.285 & 32.438 & 74.41 & \textbf{83.42} \\
Zafar et al.~\cite{zafar2024iterativeobjectcountoptimization} & 16.50 & 4.575 & 12.449 & 72.39 & 73.37  \\
CountGen~\cite{binyamin2024count} & 44.40 & 3.015 & 10.047 & 77.10 & 82.41 \\
Ours & \textbf{50.00} & \textbf{2.330} & \textbf{7.890} & \textbf{77.44} & 82.41
\\ \bottomrule
\end{tabular}
\caption{Experimantal Results on
Comparison of object count accuracy on T2I-CompBench Count ~\cite{zafar2024iterativeobjectcountoptimization} between CountCluster (Ours) and existing methods.}
\label{tab_T2I_1}
\end{table*}

\section{Comparison of Clustering Methods for Target Map Extraction}
\cref{tab_GMM1} and \cref{tab_GMM2} presents an ablation study that evaluates the effectiveness of the proposed CCL-based clustering strategy by comparing it with Gaussian Mixture Model (GMM) ~\cite{Reynolds2009} clustering.
It aims to verify whether explicitly leveraging the spatial connectivity of high-activation regions is more suitable for extracting reliable target object maps in proposed framework.

As shown in \cref{tab_GMM1}, the proposed CCL-based clustering with region count adjustment consistently achieves higher counting accuracy and better image quality than GMM-based clustering. We observe that GMM tends to form clusters around small, artifact-like local activations in the masked attention map, which do not correspond to meaningful object instances. As a result, such spurious clusters degrade both object count accuracy and overall image quality.

As shown in \cref{tab_GMM2}, While GMM-based clustering yields a slightly higher CLIPScore, our method achieves substantially higher ImageReward scores, indicating that the generated images are more preferred in terms of perceptual quality and visual coherence. These results suggest that explicitly leveraging the spatial connectivity of high-activation regions and filtering out artifact-level responses is crucial for reliable object count control.

\section{Analysis of Potential Concerns}
We construct a 2D Gaussian–based target object map and use it to guide the object CAM at the first denoising timestep to form $k$ clearly separated cluster regions.
Under this design, the following concerns may arise:
(1) Can the model reliably generate objects that are closely placed or occluded?
(2) Can it properly generate objects whose shapes are not Gaussian?
\Cref{figure8} shows that our method successfully addresses these scenarios.

Our method enforces alignment between the object CAM and the target object map only at the first denoising timestep, not across all timesteps.
This design is motivated by prior observations~\cite{cao2025temporal, 10.1016/j.eswa.2024.123231} that core concepts of objects such as a cat’s ears or a human’s head are determined during the earliest diffusion steps.
Leveraging this property, we constrain the core concept to appear only at the first timestep, while subsequent timesteps rely on the diffusion model’s inherent representational ability to generate natural object structures.
It allows our method to produce high-quality images even when objects are close to each other, occluded, or have shapes that are not Gaussian.

\begin{table}[t]
\centering
\footnotesize
\begin{tabular}{l|cc}
\toprule 
\multirow{2}{*}{Method} & \multicolumn{2}{c}{Image Quality} \\ \cline{2-3} \rule{0pt}{1.2EM}
  & CLIPScore ($\uparrow$) & ImageReward ($\uparrow$)
\\ \midrule
\multicolumn{3}{c}{SD2.1~\cite{Rombach_2022_CVPR}}
\\ \midrule
Naive SD2.1~\cite{Rombach_2022_CVPR} & 0.247 & 0.434
\\
Counting Guidance~\cite{kang2025counting} & 0.248 & \textbf{0.532}
\\
Ours & \textbf{0.248} & 0.487
\\ \midrule
\multicolumn{3}{c}{SDXL~\cite{podell2024sdxl}}
\\ \midrule
Naive SDXL~\cite{podell2024sdxl} & 0.251 & 0.379
\\
Zafar et al.~\cite{zafar2024iterativeobjectcountoptimization} & 0.239 & \textbf{0.532}
\\
CountGen~\cite{binyamin2024count} & 0.252  & 0.301
\\
Ours & \textbf{0.255} & 0.368
\\  \bottomrule
\end{tabular}
\caption{Comparison of image quality using CLIPScore~\cite{hessel2022clipscorereferencefreeevaluationmetric} and ImageReward~\cite{xu2023imagerewardlearningevaluatinghuman}.}
\vspace{-0.1cm}
\label{tab_T2I_2}
\end{table}

\section{Quantitative Results on T2I-CompBench Count}
\cref{tab_T2I_1} and \cref{tab_T2I_2} shows the experimental results on the T2I-CompBench Count~\cite{binyamin2024count} prompt set.
Compared to the integrated benchmark prompt set, this benchmark consists of prompts that specify both object counts and more complex scene compositions, making it closer to real-world scenarios.
Experimental results show that the proposed method achieves the highest object count accuracy under both CountGD- and TIFA-based evaluations on this prompt set.

In terms of image quality, the SD2-based variant of our method outperforms its corresponding base model, whereas the SDXL-based variant exhibits slightly lower image quality than the base model, indicating a trade-off between counting accuracy and image quality.
Notably, in this prompt set, higher ImageReward scores are often obtained when the complex object compositions specified in the text prompts are faithfully reflected in the generated images.
Under this condition, our method demonstrates the ability to satisfy object composition constraints while maintaining high object count accuracy.

\section{More Qualitative Results}
\begin{figure*}[t]
\centering
\includegraphics[width=1\linewidth]{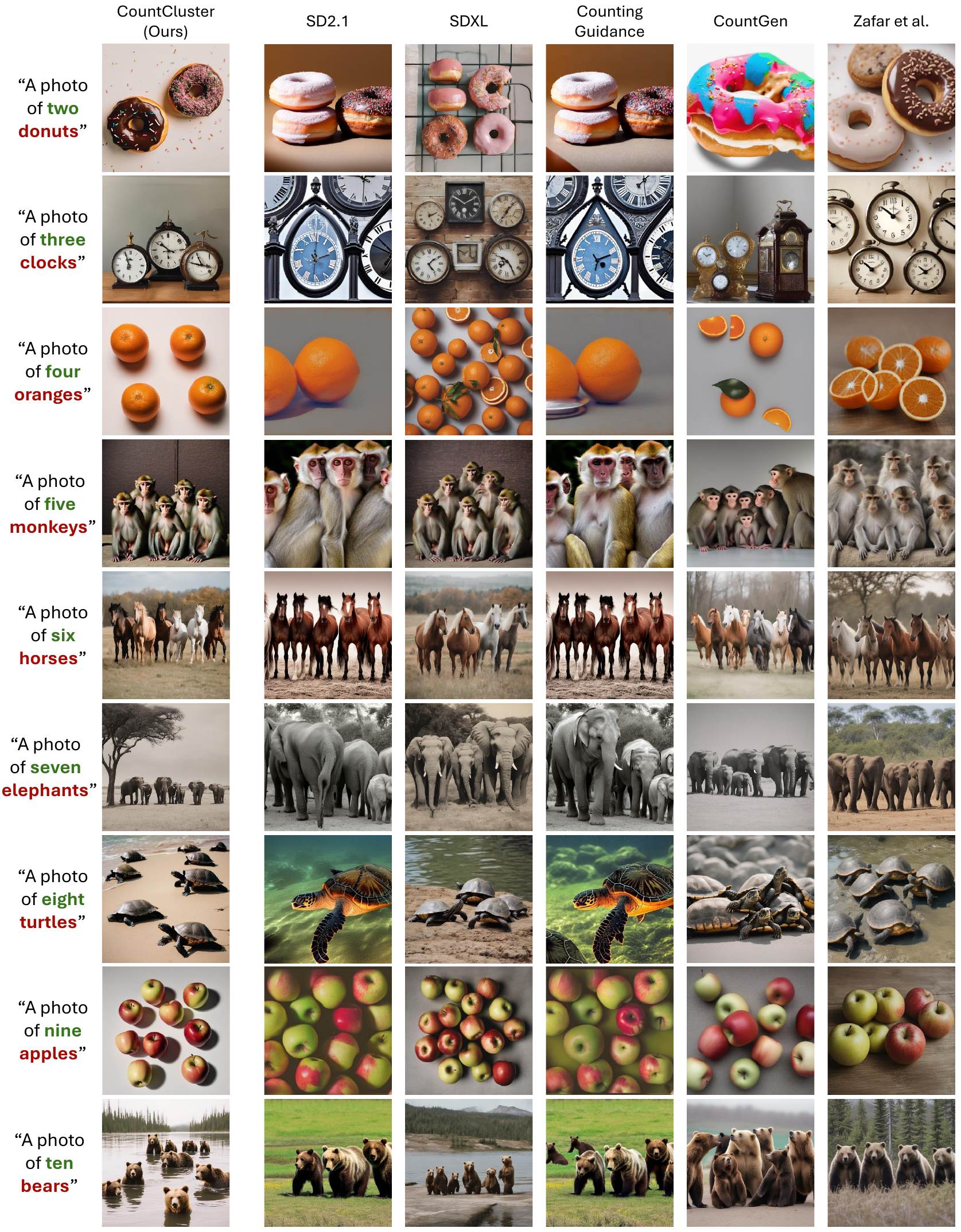}
\caption{More qualitative comparisons between CountCluster and existing methods.}
\label{figure9}
\end{figure*}

We provide additional qualitative results across a wider range of object counts and prompts. 
As shown in \cref{figure9}, existing methods often fail to match the intended number or generate instances that are not clearly distinguishable. In contrast, our method generates the correct number of instances, and the individual objects are clearly distinguishable.

\newpage

\end{document}